\def\year{2020}\relax
\documentclass[letterpaper]{article} 
\pdfoutput=1
\usepackage{aaai20}  
\usepackage{times}  
\usepackage{helvet} 
\usepackage{courier}  
\usepackage[hyphens]{url}  
\usepackage{graphicx} 
\usepackage{graphicx}  
\nocopyright
\title{Versatile Verification of Tree Ensembles}

\author{Laurens Devos, Wannes Meert, Jesse Davis\\
firstname.lastname@cs.kuleuven.be \\
Department of Computer Science,\\
KU Leuven, Leuven, Belgium}

\usepackage{bm}
\usepackage{xcolor}
\usepackage{amsmath}
\usepackage{amsthm}
\usepackage{booktabs}
\usepackage{array}

\usepackage[switch]{lineno}  %

\newtheorem{theorem}{Theorem}
\newtheorem{lemma}{Lemma}

\newcommand{\ubar}[1]{\text{\b{$#1$}}}

\newcommand{\merge}[0]{\textsc{Merge}}
\newcommand{\ouralg}[0]{\textsc{Veritas}}

\begin{document}

\maketitle

\begin{abstract}
    Machine learned models often must abide by certain requirements (e.g., fairness or legal). This has spurred interested in developing approaches that can provably verify whether a model satisfies certain properties.
    This paper introduces a generic algorithm called \ouralg{} that enables tackling multiple different verification tasks for tree ensemble models like random forests (RFs) and gradient boosting decision trees (GBDTs). This generality contrasts with previous work, which has focused exclusively on either adversarial example generation or robustness checking. \ouralg{} formulates the verification task as a generic optimization problem and introduces a novel search space representation. 
    \ouralg{} offers two key advantages.
    First, it provides anytime lower and upper bounds when the optimization problem cannot be solved exactly. In contrast, many existing methods have focused on exact solutions and are thus limited by the verification problem being NP-complete.
    Second, \ouralg{} produces full (bounded suboptimal) solutions that can be used to generate concrete examples.
    We experimentally show that \ouralg{} outperforms the previous state of the art by
    (a) generating exact solutions more frequently,
    (b) producing tighter bounds when (a) is not possible, and
    (c) offering orders of magnitude speed ups.
    Subsequently, \ouralg{} enables tackling more and larger real-world verification scenarios.
\end{abstract}


\section{Introduction}
Currently, it is becoming more common for deployed machine learned models to conform to requirements (e.g., legal) or exhibit specific properties (e.g., fairness). This has motivated the development of verification approaches that are applicable to learned models. Given a specific property, these techniques verify, that is, prove whether or not the property holds. Some examples of verification questions are:
\begin{itemize}
    \item \textbf{Adversarial example generation:} Given a data example, can slightly perturbing it cause its predicted label to flip? \cite{szegedy13,goodfellow14,einziger19}
    \item \textbf{Robustness checking:} Given a data example, what is the minimum distance to such an adversarial example? \cite{carlini2017,ranzato20}
    \item \textbf{Feature dominance:} Given a set of constraints describing a class of data examples of interest, can we find one or more attributes where changing their values would disproportionately affect the model's prediction?
    \item \textbf{Fairness:} Do two instances exist that differ only their protected attributes (e.g. sex, race, age) or a proxy variable, but have different predicted labels? \cite{dwork12}
\end{itemize}
If a property is violated, it is desirable
to also return constructed counter examples. For instance, if an adversarial example exists it is useful to see one concrete instance as this could, for example, be added to the training set. 


The popularity of additive tree ensembles (e.g., random forests~\cite{breiman01} and gradient boosting tree models~\cite{chen16,ke17,devos19}) has motivated interest in developing verification techniques for this model class. Most current work is characterized by its focus on (1) solving one specific type of verification problem (e.g., adversarial example generation~\cite{einziger19} or robustness checking~\cite{hchen19,tornblom2020,ranzato20}) and (2) developing exact solutions. While exact solutions are desirable, in practice they are not always feasible because verification tasks on additive ensembles are NP-complete~\cite{kantchelian16}. 

Recently, \citeauthor{hchen19} (\citeyear{hchen19}) took a first step towards overcoming (2) by developing a graph-based approach that computes an upper bound rather than a full solution if time and space limits are exceeded.
However, this approach has three important limitations. First, the search steps are very coarse-grained, and the algorithm often terminates after only a few steps. 
Second, the algorithm is not guided by a heuristic, so a selected step may not improve the bound. Practically, this translates into the algorithm generating looser bounds.  Third, the approach only is able to generate an example if the search terminates. 

This paper introduces a generic anytime algorithm called \ouralg{} -- \textbf{Veri}fication of \textbf{T}rees using \textbf{A}nytime \textbf{S}earch -- for addressing multiple different types of verification task on tree ensembles. This contrast with existing work that focuses on a single verification task. The key insight underlying \ouralg{} is that a large class of verification problems can be posed as a generic constrained optimization problem.  We propose a novel search space for which an admissible heuristic exists, which confers two advantages.  First, it provides anytime upper and lower bounds. Second, it can generate a concrete example using the current lower bound. Empirically, we evaluate our approach on a number of verification tasks and show that we outperform the state of art by (1)  generating exact solutions more frequently, (2) producing tighter bounds when no exact solution is found (3) offering orders of magnitude speed ups.

\section{Preliminaries}

\subsection{Additive Tree Ensembles}

The framework described in this paper reasons about \textbf{additive ensembles of binary trees}.\footnote{Note that all trees can be represented as binary trees.}
A binary tree $T$ consists of two types of nodes.
An \textit{internal node} stores a split condition and references to a left and a right child node. The split condition is a less-than comparison $X < \tau$ defined on an attribute $X$ for some threshold $\tau$. Splits on binary attributes are possible using $\tau = 0.5$.
A \textit{leaf node} has no children and simply stores an output value $\nu$ called the \textbf{leaf value}.
The \textit{root node} is the only node that has no parent.

Given a data example $\bm{x}$ from the input space $\mathcal{X}$, a tree is evaluated by recursively traversing it starting from the root node. For internal nodes, the node's split condition is tested on $\bm{x}$; if the test succeeds, the procedure is recursively applied to the left child node, else, it is applied to the right child node. If a leaf node is encountered, the leaf value is returned and the procedure terminates.

The \textbf{box} of leaf $l$, denoted $\mathrm{box}(l)$, defines a hypercube of the input space by conjoining all split condition from the root to $l$. All data examples $\bm{x}\in \mathrm{box}(l)$ evaluate to leaf $l$. For example, $\mathrm{box}(l_2^2)$ in Figure~\ref{fig:example} equals $\{\textrm{Age} < 50, \textrm{Height} \geq 200 \}$.
We also extend the definition of box to a set of leafs: $\mathrm{box}(l^1,\ldots,l^M) = \bigcap_m \mathrm{box}(l^m)$.
Two leafs \textbf{overlap} when the intersection of their boxes is non-empty. For example, in Figure~\ref{fig:example}, leaf $l^2_2$ overlaps with $l^1_1$, but does not with $l^3_3$.


An additive ensemble of trees is a sum of trees $\bm{T} = T^1 + \cdots + T^M$.
We use $l^m_i$ to denote the $i$th leaf of tree $T^m$.
Given a data example $\bm{x}$, $\bm{T}$ evaluates to the sum of the evaluations of all trees.
The set of leaf nodes whose leaf values contribute to the output of the ensemble is called an \textbf{output configuration} of the ensemble.
The box of an output configuration is by definition non-empty.
For all data examples $\bm{x}$ in an output configuration's box, the output of the ensemble is fixed because the same leafs are activated each time.
For example, in Figure~\ref{fig:example}, $\mathrm{box}(l^1_1, l^2_1, l^3_1) = \{ \textrm{Age} < 40, \textrm{Height} < 200, \textrm{BMI} < 28 \}$. All data examples in that box will evaluate to $\nu^1_1+\nu^2_1+\nu^3_1$.

\subsection{\merge{}: Robustness Verification}

\citeauthor{hchen19} (\citeyear{hchen19}) present a method for robustness verification of tree ensembles. Their algorithm produces a lower bound $\ubar{\delta}$ on the distance to the closest adversarial example for a particular example $\bm{x}$ and uses a 10-step binary search.
Given a model $\bm{T}$, a data example $\bm{x}$ with predicted label $\bm{T}(\bm{x})$,
and a maximum distance $\delta$, they verify whether an example $\tilde{\bm{x}}$ in the area of $\bm{x}$, ${||\bm{x}-\tilde{\bm{x}}||}_\infty < \delta$, exists that flips the predicted label: $\bm{T}(\tilde{\bm{x}}) \neq \bm{T}(\bm{x})$.
If such an example exists, the area surrounding $\bm{x}$ is shrunk by decreasing $\delta$, else, the area is expanded by increasing $\delta$. This is repeated 10 times or until enough precision on $\ubar{\delta}$ is obtained.

To prove that no $\tilde{\bm{x}}$ exists with a flipped label within a distance $\delta$ from $\bm{x}$, \citeauthor{hchen19} use a
graph representation of the ensemble model and a \textit{merge} procedure defined on the independent sets\footnote{An independent set is a subset of vertexes such that no two vertexes in the set are connected by an edge.} of the graph to compute an upper bound on the ensemble's output. When the upper bound is less than zero, it is impossible for the ensemble to output the positive class for any example in the area surrounding $\bm{x}$. 

\begin{figure}
    \centering
    \includegraphics{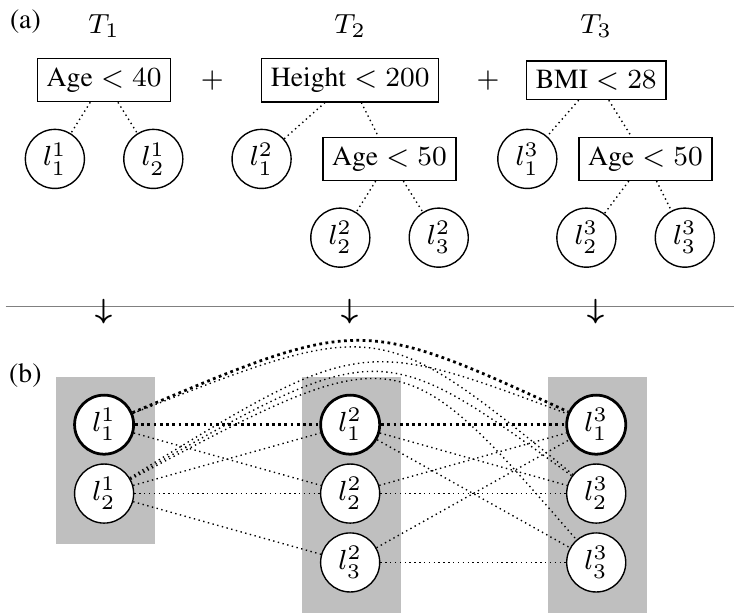}
    \caption{An example ensemble with (a) three trees and (b) its multipartite graph transformation $\bm{G}$, with $l_i^m$ the $i$th leaf of tree $T_m$. The graph consists of $M=3$ independent sets, one for each tree, and are denoted by a gray rectangular background. Edges in $\bm{G}$ connect leafs that have overlapping boxes. A max-clique contains exactly one vertex from each independent set (tree), and corresponds to an output configuration of the ensemble.
    An example max-clique in $\bm{G}$ is $[l^1_1, l^2_1, l^3_1]$ denoted by bold vertexes and bold edges.
    }
    \label{fig:example}
\end{figure}

The graph $\bm{G}$ has one vertex for each leaf $l^m_i$ of each tree $T^m$ in ensemble $\bm{T}$.
Two vertexes are connected by an edge when their boxes overlap.
Figure~\ref{fig:example} shows an ensemble and its corresponding graph representation.
When verifying robustness, only the neighborhood surrounding an example $\bm{x}$ is considered.
This is accomplished by including only the leafs that are accessible by examples close to $\bm{x}$.
Specifically, only leafs $l^m_i$ for which a $\tilde{\bm{x}} \in \mathrm{box}(l^m_i), {||\bm{x}-\tilde{\bm{x}}||}_\infty< \delta$ exists are included in the graph.
For example in Figure~\ref{fig:example}, with $\delta=5$ and $\bm{x}=\{\textrm{Age}\!\!: 60, \textrm{Height}\!\!: 180, \textrm{BMI}\!\!: 21\}$, only leafs $l^1_2$, $l^2_1$, and $l^3_1$ are included.

The graph has two key properties: (1) there is a one-to-one correspondence between the trees in the ensemble and the independent sets, and (2) there is a one-to-one correspondence between an output configuration and a max-clique.
\begin{lemma}[Lemma 1, proof in \cite{hchen19}]
    \label{lemma:hchen}
    A set of leafs is a max-clique in $G$ iff it is an output configuration.
\end{lemma}
\citeauthor{hchen19}'s main contribution is realizing that \textbf{merging} independent sets
maintains the property in Lemma~\ref{lemma:hchen} and improves the upper bound on the ensemble's output $\bar{b}$: 
\begin{equation}
    \label{eq:upper-bnd}
    \sum_{m} \max_{i} \nu^m_i,
\end{equation}
\noindent where $m$ ranges over the remaining independent sets and $\nu^m_i$ are the leaf values of the vertexes in independent set $m$.


The \merge{} algorithm has three major drawbacks. First, it is coarse-grained: the number of steps is at most $\lceil \log_2(M) \rceil$. The computational difficulty of each step grows exponentially, so the algorithm stalls after only a few steps. Second, it is not guided by a heuristic and most of its work does not improve the bound in Equation~\ref{eq:upper-bnd}. Third, the algorithm only produces a full solution when it runs to completion, i.e., all independent sets are merged into one.



\section{\ouralg{}: Verification as a Constrained Optimization Problem}

Consider the following optimization problems:
\begin{description}
        
    \item[Robustness checking] 
         Let $\bm{T}$ be a binary classifier that classifies example $\bm{x}$ as negative. 
        If $\max_{{||\bm{x}-\tilde{\bm{x}}||}_{\infty} < \delta} T(\tilde{\bm{x}}) < 0$, then no positively classified perturbed example can exist within distance $\delta$ from $\bm{x}$. 

    \item[Fairness] Suppose an insurance company estimates its clients' health scores using model $\bm{T}$ and  wants to know which attributes affect the health score of middle-aged
        women the most. 
        Given two middle-aged female individuals $\bm{x}_1$ and $\bm{x}_2$ that differ only in a single attribute, the most dominant attribute is the one that maximizes $\bm{T}(\bm{x}_2) - \bm{T}(\bm{x}_1)$. 
        The insurance company could take precautions if the most dominant attribute is a protected attribute, because that could be considered unfair.
\end{description}


Both these example belong to a class of verification problems that can be modeled as \textbf{a generic optimization problem}:
\textit{find two examples that (1) satisfy the given constraints and (2) maximize the difference between the outputs of two models}.  Formally, the optimization problem is: 
\begin{equation}
    \label{eq:opt2}
    \max_{
        \bm{x}_1, \bm{x}_2 \in \mathcal{X}
    } \bm{T}_2(\bm{x}_2)  - \bm{T}_1(\bm{x}_1)
    \quad \text{subject to}
    \quad \mathcal{C}(\bm{x}_1, \bm{x}_2).
\end{equation}
\noindent where  $\bm{T}_1$ and $\bm{T}_2$ are models, $\bm{x}_1$ and $\bm{x}_2$ are examples from the input space $\mathcal{X}$, and  $\mathcal{C} : \mathcal{X}^2 \rightarrow \{\mathrm{true}, \mathrm{false}\}$ is a function defining constraints on the input space. 
In many cases, we only consider a single model $\bm{T}$. Taking $\bm{T}_1$ to be the trivial model that always predicts 0.0, and $\bm{T}_2 = \bm{T}$, yields the maximization problem:
\begin{equation}
    \label{eq:opt1}
    \max_{\bm{x}\in\mathcal{X}} \bm{T}(\bm{x}) 
    \quad \text{subject to}
    \quad \mathcal{C}(\bm{x}).
\end{equation}
Minimizing $\bm{T}$ is also possible by taking $\bm{T}_1 = \bm{T}$, and making $\bm{T}_2$ the trivial model.
Note that we simplify $\mathcal{C}$ because the example of the trivial model does not affect the outcome.

The algorithm presented in this paper, \ouralg{}, is a search algorithm that solves the optimization problem in Equation~\ref{eq:opt2}.
The algorithm operates in a novel search space derived from \citeauthor{hchen19}'s graph representation $\bm{G}$. It is fine-grained as a single step is cheap, heuristically guided and does not waste time in irrelevant parts of the search space.
The algorithm is anytime and outputs ever-improving upper bounds, lower bounds, and suboptimal full solutions as it is running. As more time and memory is provided, the bounds grow closer until the exact full solution is found.
As this problem is NP-complete \cite{kantchelian16}, it might take a long time until the bounds converge to an exact solution. This makes the anytime nature of \ouralg{} particularly compelling.

We will first introduce the search space. Then, we will show how this space is used to solve the optimization problem in Equation~\ref{eq:opt2}. For ease of explanation, we will first tackle the single-instance setting in Equation~\ref{eq:opt1}, and then extend our method to support the two-instance setting.

\subsection{The Search Space $S$}


The search is not executed directly in $\bm{G}$, but rather in a \textbf{sound and complete} search space $S$ derived from $\bm{G}$.
The states of $S$ are cliques in $\bm{G}$ and are represented as sequences of leaf nodes.
Starting from the initial empty sequence $[\,]$ at depth 0, the search space recursively builds up output configurations by adding leafs to the state in order, one leaf per tree.
Specifically, a state $s = [l^1_{i_1}, \ldots, l^m_{i_m}]$ at depth $m$ is expanded to states $C(s)$ at depth $m+1$:
%
%
\begin{align}
    &C([l^1_{i_1}, \ldots, l^m_{i_m}]) = \nonumber\\
    &\qquad
    \{
    [l^1_{i_1}, \ldots, l^m_{i_m}, l^{m+1}] \mid
    l^{m+1} \in L^{m+1},\nonumber\\
    &\qquad\qquad
    \mathrm{box}(l^1_{i_1}, \ldots, l^m_{i_m}, l^{m+1}) \neq \emptyset
    \label{eq:children}
    \},
\end{align}
where $L^{m+1}$ is the set of all leafs of tree $T^{m+1}$.
The expand function $C$ ensures that each expanded state is again a clique in $G$ by allowing only states with non-empty boxes.
When depth $m=M$ is reached, an output configuration, or equivalently a max-clique is found.
The search space has two important properties. For brevity, we will write $\mathrm{box}(s)$ to indicate the box of the set of leafs in the search state $s\in S$.
\begin{theorem}
    \label{thrm:sound-complete}
    Each state at depth $m=M$ corresponds to an output configuration (i.e., a max-clique in $\bm{G}$), and the search space enumerates all output configurations, that is, the enumeration is \textit{sound} and \textit{complete}.
\end{theorem}
\begin{proof}
    We show that $S$ is \textit{sound} and \textit{complete}.

            \textit{Soundness:}
            We show that any state $s=[l^1_{i_1}, \ldots, l^M_{i_M}]$ is an output configuration.
            By definition of $C$ in Equation~\ref{eq:children}, the box of the state is non-empty.
            For each data example $\bm{x}$ in the state's box, it holds that $\bm{x} \in \mathrm{box}(l^{m}_{i_{m}})$, $m = 1,\ldots,M$. Because of the properties of tree evaluation and the definition of $\mathrm{box}$, tree $T^m$ evaluates $\bm{x}$ to leaf $l^m_{i_m}$. Therefore, a data example $\bm{x}$ exists that activates all leaf nodes in $s$, and the configuration is valid.
            Using Lemma~\ref{lemma:hchen}, it follows that $s$ is a max-clique in $\bm{G}$.

            \textit{Completeness:}
            Assume an output configuration $O=\{l^1_{i_1}, \ldots, l^M_{i_M}\}$ so that $[l^1_{i_1}, \ldots, l^M_{i_M}]$ is not a state in the search space.
            Let $s=[l^1_{i_1}, \ldots, l^m_{i_m}] \subset O$ and $t=[l^1_{i_1}, \ldots, l^m_{i_m}, l^{m+1}_{i_{m+1}}] \subseteq O$ be sequences of leaf nodes such that $s$ is a state in $S$, but $t$ is not. It must be that $t \notin C(s)$.
            This can only be true when $\mathrm{box}(t) = \mathrm{box}(l^1_{i_1}, \ldots, l^m_{i_m}, l^{m+1}_{i_{m+1}}) = \bigcap_{m'=1}^{m+1} \mathrm{box}(l^{m'}_{i_{m'}}) = \emptyset$. 
            This contradicts the fact that $O$ is an output configuration.
\end{proof}


\subsection{Finding The Best State at Depth $M$}

To solve the optimization problem in Equation~\ref{eq:opt1}, we need to find the state in $S$ at depth $M$ with the maximum output that adheres to the constraints $\mathcal{C}$.
Constructing $S$ explicitly and choosing the optimal state by enumerating all possibilities is intractable in practice.
For that reason, we introduce an \textbf{admissible and consistent} heuristic to traverse the space in a best-first fashion. This allows us to only materialize the parts of $S$ that are relevant to the problem.


The search is based on A*. It maintains an OPEN list of states in $S$ and repeatedly \textit{expands} the state in OPEN with the best $f$-value until a state at depth $M$ is found. It removes the current best state $s=[l^1_{i_1},\ldots,l^m_{i_m}]$ from OPEN and adds all its successors $C(s)$ to OPEN. The $f$-value is the sum
of two values: $g(s)$, the sum of the leaf values of the leafs in the state, and the heuristic $h(s)$, an estimation of the remaining value to a state at depth $M$:
\begin{align}
    \label{eq:g}
    g(s) &= \sum_{m'=1}^m \nu^{m'}_{i_{m'}},\\
    \label{eq:h}
    h(s) &= \sum_{m'=m+1}^M h_{m'}(s), \\
    \label{eq:hm}
    h_{m'}(s) &= 
    \max\{
        \begin{array}[t]{l}
            \nu^{m'}_j
            \mid
            l^{m'} \in L^{m'},\\
            \mathrm{box}(l^1_{i_1},\ldots,l^m_{i_m}, l^{m'}) \neq \emptyset
            \ \}.
        \end{array}
\end{align}
Intuitively, the heuristic $h(s)$ sums upper bounds $h_{m'}(s)$ for all remaining trees $T^{m'}$ for which no leaf has been added to the state yet. The upper bound $h_{m'}(s)$ is the maximum leaf value of any leaf of $T^{m'}$ that overlaps with all leafs in $s$.
Note that we do not need to keep a VISITED list, because there are no cycles. We show that this heuristic leads us to the optimal solution next.

\begin{theorem}
    \label{thrm:optimal}
    The heuristic search in $S$ is guaranteed to find the max-clique in $\bm{G}$ which corresponds to the optimal output configuration of the optimization problem in Equation~\ref{eq:opt1}.
\end{theorem}


\begin{proof}
    We need to show that the heuristic is admissible and consistent \cite{pearl85}.

        \textit{Admissible:}
            For maximization problems, a heuristic is admissible when it never underestimates the remaining value to the goal state.
            This is the case because each leaf considered for extension in Equation~\ref{eq:children} is also considered in the $\max$ of Equation~\ref{eq:hm}.

        \textit{Consistent:}
            For each state $s=[l^1_{i_1},\ldots,l^m_{i_m}]$ and any extension $t = [l^1_{i_1},\ldots,l^m_{i_m},l^{m+1}_j]$, $h$ is consistent when $h(s) \geq \nu_j^{m+1} + h(t)$.
            Unlike $h(t)$, $h(s)$ contains a term $h_{m+1}(s)$, which is an overestimation of $\nu_j^{m+1}$, that is, $h_{m+1}(s) \geq \nu_j^{m+1}$. Additionally, each $h_{m'}(s) \geq h_{m'}(t), m+1 < m' \leq M$ because
            $\mathrm{box}(s) \supseteq \mathrm{box}(t)$, and consequently, each $\nu_j^{m'}$ considered in the $\max$ of $h_{m'}(t)$ is also considered in the $\max$ of $h_{m'}(s)$ (Equation~\ref{eq:hm}). 
\end{proof}



\subsection{Anytime Upper and Lower Bound Estimates}

Depending on the size and difficulty of the problem, the search defined above might take a long time to find a solution. However, even when the search is terminated prematurely, the current best state in the OPEN list still contains useful information: because $h$ is admissible, the current best $f$-value is an \textbf{upper bound} $\bar{b}$ on the optimal output.

To also produce a lower bound on the maximum output of the ensemble, we borrow ideas from Anytime Repairing A* (ARA*) \cite{likhachev2004}. The importance of the heuristic in the $f$-value is reduced by $\epsilon$, $0 < \epsilon \leq 1$ as follows:
\begin{equation}
    \label{eq:f-ara}
    f(s) = g(s) + \epsilon h(s).
\end{equation}
This promotes deeper solutions, yielding full solutions much quicker.
However, when $\epsilon < 1$, the solutions might no longer be optimal because $\epsilon h(s)$ is no longer an admissible heuristic. Nonetheless, the suboptimality is bounded by $\epsilon$: for a suboptimal solution $\tilde{\bm{x}}$, the optimal solution's output is no larger than $\bm{T}(\tilde{\bm{x}})/\epsilon$.
A suboptimal solution $\tilde{\bm{x}}$ is at least as good as the optimal solution, so $\bm{T}(\tilde{\bm{x}})$ is a \textbf{lower bound} $\ubar{b}$ on the optimal output.


Both the upper bound and the lower bound are anytime: The best $f$-value in the OPEN list is always accessible. The $\epsilon$ value in the relaxed $f$-score of Equation~\ref{eq:f-ara} can be gradually increased each time a suboptimal solution is found. Using ARA* means that it is not necessary to restart the search from scratch: the OPEN list can be reused, making the incremental increase of $\epsilon$ a cheap operation.\footnote{Our case is more simple than the general case in \cite{likhachev2004} because we do not have cycles.}

\subsection{Incorporating Constraints into the Search} 

A naive way of incorporating the constraint function $\mathcal{C}$ in Equations~\ref{eq:opt2} and \ref{eq:opt1} is to ignore it during the search, and filter solutions returned by the search. This is inefficient, as the constraints often rule out large chunks of the search space.

For that reason, we lift $\mathcal{C}$ to the search state level.
For example, considering the example in Figure~\ref{fig:example} and the constraint $\textrm{Age} > 60$, we reject states $[l^1_1]$, $[l^1_2, l^2_2]$, and $[l^1_2, l^2_1, l^3_2]$.

Let $\mathcal{C}_s : S \rightarrow \{ \mathrm{true}, \mathrm{false} \}$ be a function mapping the states of the search space to accept or reject. The function has the following two properties: 
\begin{enumerate}
    \item It is consistent with $\mathcal{C}$: a state $s$ is accepted by $\mathcal{C}_s$ if and only if it is possible to find a data example $\bm{x}$ that sorts to the leafs in $s$ and $\mathcal{C}(\bm{x})$ is true.
\item It is consistent across states: if $\mathcal{C}_s(s) = \mathrm{false}$, and $t \supseteq s$ is a descendant from $s$, then $\mathcal{C}_s(t) = \mathrm{false}$. In words: if a state is rejected by $\mathcal{C}_s$, then all its state expansions must also be rejected. This property ensures that we do not reject states that are the predecessor of a valid state.
\end{enumerate}
Because all split conditions in the tree ensemble are simple linear constraints, the two properties are trivially satisfied.
The constraint function $\mathcal{C}_s$ can be integrated into the search procedure by only adding accepted states to the OPEN list. For simple linear constraints, we can prune the graph $\bm{G}$ before the search starts, just like in the \merge{} algorithm.

\subsection{Maximizing the Difference Between Two Models}

Now we explain the necessary adjustments to enable solving Equation~\ref{eq:opt2}.
The search space is modified as follows:
a state is extended to consist of two sequences of leafs, one sequence for each instance. The expand function alternately expands the first and the second instance using Equation~\ref{eq:children}.
The $f$-value maximized by the search is updated to the difference between the $f$-values of the two instances:
\begin{align}
    f(s_1, s_2) &= f_2(s_2) - f_1(s_1)\nonumber\\
    &= g(s_2) - g(s_1) + h_2(s_2) - h_1(s_1).
\end{align}
The $g$ (Equation~\ref{eq:g}) and $h_2$ (Equation~\ref{eq:h}) remain unchanged. The first heuristic $h_1$ is an admissible and consistent heuristic with respect to the minimizing variant of Equation~\ref{eq:opt1} obtained by replacing $\max$ by $\min$ in Equation~\ref{eq:hm}. This makes $h_2 - h_1$ an admissible and consistent heuristic for the optimization problem in Equation~\ref{eq:opt2} (proof in appendix).

\subsection{Summarizing the Setup}

We now have all necessary tools to answer verification questions. First, train a tree ensemble. Second, define the constraint function $\mathcal{C}_s$. The complexity of this step greatly depends on the verification question.\footnote{In general, this step can be framed as a SAT problem: is it possible that an $\bm{x} \in \mathrm{box}(s)$ exists that satisfies the constraints.} Third, give the learned model and $\mathcal{C}_s$ to \ouralg{}, which will produce a stream of decreasing upper bounds $\bar{b}$, and a stream of increasing lower bounds $\ubar{b} = \bm{T}(\tilde{\bm{x}})$, with $\tilde{\bm{x}}$ suboptimal full solutions. Given infinite time and memory, the bounds converge: $\bar{b} = \ubar{b}$, and the last $\tilde{\bm{x}}$ is the optimal solution.


\section{Experimental Evaluation}

We compare our method to \merge{} on four tasks: (1) robustness verification, \merge{}'s core task, (2) a stress test on four datasets with varying model sizes, (3) randomly generated verification questions, and (4) two new general use cases. The last task surpasses \merge{}'s capabilities.

\merge{} is the current state of the art for robustness verification.
It outperforms an exact mixed-integer linear programming (MILP) approach \cite{kantchelian16}, and a linear programming relaxation of the MILP approach. \merge{} is orders of magnitude faster than the former and offers a better bound than the latter.
We do not compare to Silva \cite{ranzato20} and VoTE \cite{tornblom2020} because their systems do not estimate the distance to the closest adversarial example. Rather, they only check robustness for $\delta=1$, the easiest, most restrictive case.
Furthermore, as \citeauthor{tornblom2020} also indicate, VoTE struggles to scale to datasets with a larger number of attributes like MNIST.

\begin{figure}
    \centering
    \small 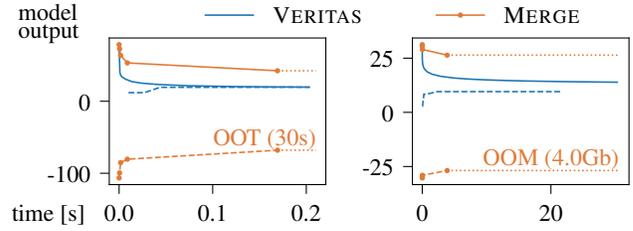
    \caption{Two examples of how the upper and lower bounds progress over time. \ouralg{} finds the exact solution on the left, and tight upper and lower bounds on the right. OOT and OOM indicate that \merge{} ran out of time and memory respectively.
    }
    \label{fig:bound-over-time}
\end{figure}
To compare the run times of \ouralg{} and \merge{}, we measure the time taken by \ouralg{} to reach \merge{}'s best bound. We refer to this as the \textit{time to bound} or \textit{TTB} statistic.
Figure~\ref{fig:bound-over-time} shows two examples of how the bounds of \ouralg{} and \merge{} improve over time.
The \textit{TTB} statistic is the time for \ouralg{}'s upper bound to reach the $y$-value of the upper dotted line in Figure~\ref{fig:bound-over-time}.

We set \merge{}'s $L$ parameter (the number of independent sets merged per step) to 2 to make \merge{} as fine-grained as possible.
We run \merge{} until either it finds the optimal solution or time or memory runs out.
Originally, \merge{} only produced an upper bound $\bar{b}$.
We use Equation~\ref{eq:upper-bnd} with $\max$ replaced by $\min$ as a lower bound for \merge{}.

We use XGBoost to train GBDT models. As we will be comparing different model sizes, we choose an appropriate tree depth and ensemble size per experiment,
and use hyperparameter optimization to find a suitable learning rate for a particular tree depth and ensemble size pair. All experiments ran on an Intel(R) Core(TM) i7-2600 CPU with 16GB of memory. Details about datasets, timeouts, memory limits, and model parameters can be found in the appendix.

\subsection{Verifying Robustness}

In this first experiment, we show how better upper bounds $\bar{b}$ on the ensemble's output improve $\ubar{\delta}$, the lower bound estimate on the distance to the closest adversarial example.
We trained a multi-classification GBDT model $\bm{T}$ with 50 trees of depth 5 on MNIST. The model consists of one binary one-vs-all classifier $\bm{T}_y$ per digit. We pick a random digit $\bm{x}$ with predicted label $y=\bm{T}(\bm{x})$ from the dataset and a target label $y'\neq y$, and use the binary search procedure to compute $\ubar{\delta}$ for the closest adversarial example labeled $y'$ instead of $y$.
The binary search uses the upper bound $\bar{b}$ on the maximal value of $\bm{T}_{y'}(\tilde{\bm{x}}) - \bm{T}_y(\tilde{\bm{x}})$, ${||\bm{x}-\tilde{\bm{x}}||}_\infty < \delta$, to direct its updates to $\delta$.
It starts at $\delta=20$ and increases $\delta$ when no adversarial example exists ($\bar{b} <0$), or decreases $\delta$ when an adversarial example \textit{may} exist ($\bar{b} \geq 0$). The search takes 10 such steps and ultimately produces a lower bound $\ubar{\delta}$.
This procedure is repeated 1260 times, yielding 12,600 upper bound $\bar{b}$ calculations. The performance of \ouralg{} and \merge{} is summarized below:

\begin{enumerate}
    \item \ouralg{} generates a better upper bound $\bar{b}$ in all cases. This leads to a divergence in the binary search in 15.5\% of the cases, resulting in better lower bounds $\ubar{\delta}$. In all other cases, \ouralg{} and \merge{} produce the same $\ubar{\delta}$.
    \item \ouralg{} can prove that its $\ubar{\delta}$ value is optimal in 99\% of the cases. It does this by generating an adversarial example (a full solution) for a $\delta_1$, and proving that no adversarial example for a $\delta_2$ exists, and $\lfloor \delta_1 \rfloor = \lceil \delta_2 \rceil$. Because MNIST attribute values are integers, the minimal distance is the integer that has $\delta_1$ and $\delta_2$ on either side.
    \item \ouralg{} is faster than \merge{} for 98\% of the $\bar{b}$ calculations using the \textit{TTB} statistic. In 59\% of the cases, our algorithm is more than 1000 times faster.
\end{enumerate}
\begin{figure}
    \centering
    \small 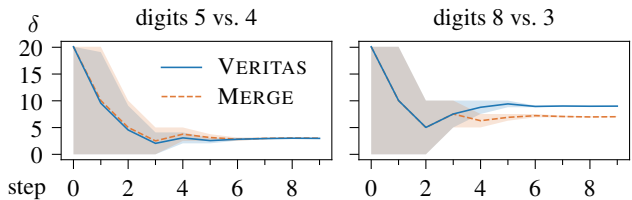
    \caption{Illustrations of two binary search executions on two MNIST digits, a 5 and an 8. \ouralg{} and \merge{} compute a lower bound $\ubar{\delta}$ on the distance to the closest adversarial example that is classified as a 4 (left) and a 3 (right).
    The background colors indicate the remaining binary search interval on $\delta$. A higher $\ubar{\delta}$ value is better.}
    \label{fig:robust}
\end{figure}
Two example binary search executions are illustrated in Figure~\ref{fig:robust}. The left shows an example where both lower bounds $\ubar{\delta}$ converge. The right shows an example where \ouralg{} outputs a better $\ubar{\delta}$ value.

\subsection{Stress Testing}

\begin{figure}
    \centering
    \small 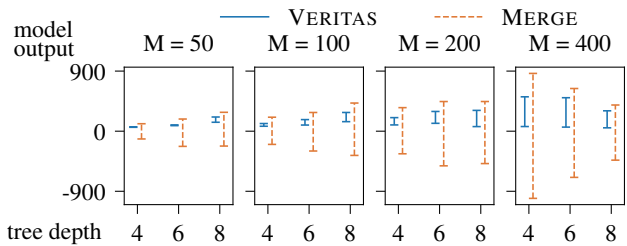
    \caption{Comparison of bounds produced by \ouralg{} and \merge{} for a varying number of tree depths and ensemble sizes on Covertype.}
    \label{fig:unconstrained}
\end{figure}

Robustness verification is inherently strongly constrained: it only considers a small neighborhood surrounding a particular fully defined example. This greatly reduces the complexity of the verification problem.
In terms of the number of possible solutions, the most difficult verification task is simply ``\textit{what is the maximal output value of the ensemble?}''  We compare \ouralg{} and \merge{} on this task for four datasets: Covertype, Allstate, California housing, and Higgs. To explore how the approaches scale on this task, we vary both the number of trees in the ensemble ($M \in \{50, 100, 200, 400\})$ and the maximum depth of the learned trees. Figure~\ref{fig:unconstrained} shows the upper and lower bounds for both methods for the Covertype dataset.
\ouralg{} finds substantially tighter bounds than \merge{} in all cases and produces usable bounds even for large and deep models. The same holds for the other datasets (results in appendix).

\subsection{Randomly Generated Verification Tasks}

To get a better idea of how both \ouralg{} and \merge{} perform on a wide range of verification questions, we randomly generate $\tau_{\min} \leq X < \tau_{\max}$ constraints on the input attributes of the Allstate, California housing, Covertype and Higgs datasets. We then compute the maximum output of an ensemble trained on a particular dataset
given the generated constraints.
Recall that a leaf is unreachable given a constraint on an input attribute when the constraint conflicts with the leaf's box. For example, in Figure~\ref{fig:example}, the constraint $\textrm{Age} < 55$ makes $l^1_2$, $l^2_3$, and $l^3_3$ unreachable.
We tune the difficulty of a problem by varying the number of accessible leafs; for the easiest problems, only 10\% of the leafs are reachable, whereas for the hardest problems, (nearly) all leafs are reachable.
Reducing the number of accessible leafs ultimately reduces the number of possible output configurations, which in turn cuts down the number of possible outcomes of the search problem.
The ensembles uniformly range in size from 25 to 500 trees and in tree depth from 3 to 8.
Table~\ref{tbl:constrained} shows the aggregated results for 2160 generated problems per dataset. The upper bound of \ouralg{} is better in all cases. The lower bound is better in more than 99\% of the cases, and is worse only when the relaxed search cannot find a suboptimal full solution in time.
\begin{table}
    \centering
    \footnotesize
    \newcommand{\myrndfig}[1]{\includegraphics[page=#1]{table_random_figures.pdf}}
\begin{tabular}{m{0.4cm}m{1.1cm}m{0.9cm}m{1.1cm}m{0.9cm}m{0.9cm}}
\toprule
 & Exact & Exact & Gap & Gap & TTB \\
Data &  \ouralg{} &  \merge{} &  \ouralg{} &   \merge{} &             \\
\midrule
   A &           32.0\% &           6.2\% &   \myrndfig{1} &   \myrndfig{2} &   \myrndfig{3} \\
  CH &           95.7\% &          35.0\% &   \myrndfig{4} &   \myrndfig{5} &   \myrndfig{6} \\
  CT &           80.2\% &          16.6\% &   \myrndfig{7} &   \myrndfig{8} &   \myrndfig{9} \\
   H &           35.5\% &           8.6\% &  \myrndfig{10} &  \myrndfig{11} &  \myrndfig{12} \\
\bottomrule
\end{tabular}

    \caption{Results for randomly generated problems for Allstate (A), California housing (CH), Covertype (CT), and Higgs (H). Each row summarizes 2160 random problems. The second and third column show how often a problem is solved exactly by the two methods. The \textit{gap} columns show the relative gap distributions. The bars show the fraction of cases where the gap is $<\!\!1\%$ (blue), up to $20\%$, $50\%$, $100\%$, and $>\!\!100\%$. A smaller gap is better.
    The \textit{TTB} histogram shows the fraction of cases where \ouralg{} is slower (first bar), or faster by a factor of up to 10, 100, 1000, and $>\!\!1000$ (next 4 green bars).
    }\label{tbl:constrained}
\end{table}

Table~\ref{tbl:constrained} contains three more per-dataset statistics. The second and third columns show that \ouralg{} is capable of finding the exact optimal solution 3 to 5 times more frequently than \merge{}. To compare the quality of the bounds when the optimal solution is not found, we measure the relative \textbf{gap} size $(\bar{b}-\ubar{b})/ \bar{b}$ between the upper and the lower bounds $\bar{b}$ and $\ubar{b}$. A smaller gap indicates tighter bounds.
The gap histograms show that the bounds generated by \ouralg{} are much better than those generated by \merge{}. For the first three datasets, \ouralg{}'s gap is at most 50\% in more than 95\% of the cases (69\% for Higgs). This statistic is between 3 and 11\% for \merge{}.
As in the robustness experiment, we use the \textit{TTB} statistic to show that \ouralg{} is often orders of magnitude faster than \merge{}. Overall, our algorithm is faster in 98\% of the cases (90\% for Higgs), and in 15\% to 25\% of the cases, our algorithm is more than 1000 times faster.

\subsection{Two New Verification Use Cases}

We now present two examples of real-world verification questions 
that our approach can solve but \merge{} cannot because it lacks support for a state constraint function $\mathcal{C}_s$ and lacks the ability to generate suboptimal full solutions.

\subsubsection{Finding Dominant Attributes: YouTube}

For this experiment we use a dataset generated from trending YouTube videos.
The task is to predict the order of magnitude of views given a bag-of-words representation of the words used in the title and description of the video. We use a GBDT model with 100 trees of depth 10. Given a number of initial words, we ask \ouralg{} to produce $k$ additional words such that the predicted views count is maximized. The state function $\mathcal{C}_s$ checks the \textit{at-most-$k$} constraint. Some examples:
\begin{itemize}
    \item \textit{live, breaking, news, war} $\rightarrow$ adding words \textit{big} and \textit{trailers} increases the prediction by 2 orders of magnitude.
    \item \textit{epic, challenge} $\rightarrow$ adding words \textit{album}, \textit{video}, and \textit{remix} increases the prediction by almost 5 orders of magnitude.
\end{itemize}
This approach is very general: it is a generic strategy that allows reasoning about the importance or dominance of one or more attributes. Other examples are fairness: given a set of constraints on the input space, maximize or minimize the output of the ensemble when only varying the values of one or more protected (proxy) attributes. If \ouralg{} finds examples of individuals that are treated significantly differently, then that might indicate unwanted model behavior.

\subsubsection{Domain Specific Questions: Soccer}

\begin{figure}
    \centering
    \small 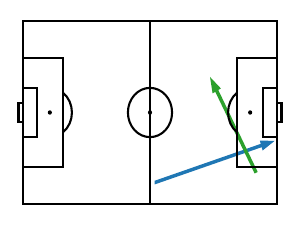
    \caption{The blue arrow shows the optimal position for a pass from the midfield to end. The green arrow shows the result for the backwards pass question: \ouralg{} generates a cut back with a positive goal probability. }
    \label{fig:soccer}
\end{figure}

For this experiment we train a model that predicts the probability of scoring a goal in soccer within the next 10 actions \cite{decroos19}. The model has 126 trees (early stopping) of depth 10. The input for the model is two consecutive game states described by the position of the ball and the two action types (pass, shot, etc.). We can ask a number of interesting domain specific questions: \textit{what ball action do we need to perform to maximize the goal probability?} or \textit{in which contexts does a backwards pass increase the probability of scoring in the next 10 actions?} Figure~\ref{fig:soccer} shows two generated instances. The state constraint function $\mathcal{C}_s$ checks two \textit{one-out-of-$k$} constraints corresponding to the one-hot-encoded action types.

\section{Related work}

A considerable amount of work has been done on verification of tree ensembles. Most work has focused on \textit{adversarial attacks}~\cite{einziger19} and robustness~\cite{kantchelian16,hchen19,ranzato20,tornblom2020}.
\citeauthor{tornblom2020} introduced the VoTE framework, a system that enumerates all \textit{equivalence classes} -- sets of data examples that evaluate to the same output value, equivalent to the concept of output configurations in this paper -- and checks whether some property holds. The properties that can be tested are general. However, the approach is limited by the number of equivalence classes, which quickly grows exponentially large~\cite{tornblom18,tornblom19,tornblom2020}.

Logical SMT theorem provers have also been used \cite{einziger19,sato19,devos20}, as have mixed-integer linear programming tools \cite{kantchelian16}. 
These approaches translate ensemble models to their respective languages and apply general purpose solvers to prove certain properties of the models.
Others have applied tools from program analysis like \textit{abstract interpretation} to verification of tree ensembles \cite{ranzato20,drews20,calzavara20arxiv}.
There is also work that focuses on learning robust models~\cite{hchen19robust,calzavara20}. Rather than verifying the robustness of existing models, these methods build models that are less susceptible to adversarial attacks.

\section{Conclusion}

We introduced \ouralg{}, a tree ensemble verification tool that is capable of solving verification tasks that can be modeled as a generic optimization problem.
It operates in a novel sound and complete search space and traverses that space using an admissible and consistent heuristic. \ouralg{} is the first anytime algorithm to produce both an upper and a lower bound on the output of a tree ensemble model.
Additionally, it also generates full suboptimal solutions that converge to the optimal solution when given enough time and memory.
We empirically show that \ouralg{} outperforms the state of the art both in terms of quality of the bounds, as well as in terms of run time.

\paragraph{Acknowledgments}

LD is supported by Research Foundation-Flanders (FWO). This research received funding from Research Foundation-Flanders under EOS No. 3099257 and KU Leuven Research Fund (C14/17/070).

\appendix

\section{Appendices}

\subsection{Proof of Consistent Heuristic}

We prove that $h_2(s_1)-h_1(s_1)$ is a consistent heuristic for the optimization problem in Equation 2 of the main paper (admissibility follows from consistency).
First, we show that $h_1$ is a consistent heuristic:

\begin{theorem}
    \label{thrm:consistent2}
    The heuristic obtained by replacing $\max$ by $\min$ in Equation~7 in the main paper, i.e.,
    \begin{equation*}
    h_1(s_1) = \sum_{m'=m+1}^M \min\{
        \begin{array}[t]{l}
            \nu^{m'} \mid
            l^{m'} \in L^{m'},\\
            \ \ \mathrm{box}(l_{i_1}^1, \ldots, l_{i_m}^m, l^{m'}) \neq \emptyset
        \ \},
        \end{array}
    \end{equation*}
    is a consistent heuristic for the minimization problem $\min_{\bm{x}_1} \bm{T}_1(\bm{x}_1)$.
\end{theorem}
\begin{proof}
    For minimization problems, consistent means that $h_1(s_1) \leq \nu + h_1(t_1)$ for two consecutive states $s_1$ and $t_1$, and the leaf value $\nu$ added to $g_1(t_1) = g_1(s_1) + \nu$.
    This proof is analogous to the proof of Theorem~2 in the main paper. 
\end{proof}

\begin{theorem}
    \label{thrm:consistent3}
    The heuristic $h(s_1, s_2) = h_2(s_2)-h_1(s_1)$ is a consistent heuristic for the optimization problem in Equation 2 of the main paper, i.e., 
\begin{equation*}
    \max_{
        \bm{x}_1, \bm{x}_2 \in \mathcal{X}
    } \bm{T}_2(\bm{x}_2)  - \bm{T}_1(\bm{x}_1)
    \quad \text{subject to}
    \quad \mathcal{C}(\bm{x}_1, \bm{x}_2).
\end{equation*}
\end{theorem}

\begin{proof}
    We show that $h(s_1, s_2) \geq -\nu_1 + h(t_1, s_2)$ and $h(s_1, s_2) \geq \nu_2 + h(s_1, t_2)$, for successive state pairs $(s_1, t_1)$  and $(s_2, t_2)$ and leaf values $\nu_1$ and $\nu_2$ added when transitioning from $s_1$ to $t_1$ and $s_2$ to $t_2$ respectively.

    We use the fact that $h_1$ is a consistent heuristic with respect to the minimization of $\bm{T}_1(\bm{x}_1)$ ($h_1(s_1) \leq \nu_1 + h_1(t_1)$)
    and $h_2$ is a consistent heuristic with respect to the maximization of $\bm{T}_2(\bm{x}_2)$ ($h_2(s_2) \geq \nu_2 + h_2(t_2)$).

    For the first case:
    \begin{align*}
        h(s_1, s_2) &= h_2(s_2) - h_1(s_1) \\
        &\geq h_2(s_2) - \left( \nu_1 + h_1(t_1) \right) \\
        &= -\nu_1 + h(t_1, s_2).
    \end{align*}

    For the second case:
    \begin{align*}
        h(s_1, s_2) &= h_2(s_2) - h_1(s_1) \\
        &\geq \left(\nu_2 + h_2(t_2) \right) - h_1(s_1) \\
        &= \nu_2 + h(s_1, t_2).
    \end{align*}

\end{proof}

\subsection{Experiments}
\subsubsection{Dataset Details}

We used the following datasets in our experiments.
\begin{itemize}
    \item \textbf{Allstate}: an insurance claim dataset available from \url{https://www.kaggle.com/c/allstate-claims-severity} with 116 categorical and 14 numerical features and 188k rows. Each row describes an insurance claim, and the target variable is a continuous \textit{loss}.
    \item \textbf{California housing}: a well-known dataset available from \url{https://github.com/ageron/handson-ml/tree/master/datasets/housing} with 10 attributes and 20k rows. The task is to predict the median income of a census block.
    \item \textbf{Covertype}: a forestry dataset available from \url{https://archive.ics.uci.edu/ml/datasets/covertype} with 54 attributes and 581k rows. The task is to predict one out of seven forest cover types. We turned it into a balanced binary classification task by predicting class 2 versus all.
    \item \textbf{Higgs}: a high energy physics dataset available from \url{https://www.kaggle.com/c/higgs-boson/data}. There are multiple versions of this dataset; we used the version from Kaggle. It has 33 attributes and 250k rows. The task is to classify whether the rows represents a decaying Higgs boson or not.
    \item \textbf{MNIST}: a famous 0-9 digit recognition task available from \url{http://yann.lecun.com/exdb/mnist/}. It consists of 70k 28-by-28 pixels gray scale images of handwritten digits.
    \item \textbf{YouTube}: a dataset derived from the collection of trending videos available from \url{https://www.kaggle.com/datasnaek/youtube-new} (US, UK and CA only). We trained on a bag-of-words representation of the text in the title and the description of the video. The target is the log of the total number of views of the video. The 330 binary attributes represent the presence of individual words (no $n$-grams) for 120k videos. 
    \item \textbf{Soccer}: This is the only non-public dataset. The data can be purchased from providers like Opta. We used two consecutive ball actions described by their $x$ and $y$ coordinates on the pitch, and the two corresponding action types as the input data. The target is to predict whether a goal occurs within the next 10 actions. We obtained a model by following the steps at \url{https://github.com/ML-KULeuven/socceraction}.

\end{itemize}

\subsection{Experiment Parameters}

We use the default XGBoost parameters with the exception of number of trees, tree depth, and learning rate. The learning rate is always determined using hyper-parameter optimization given an ensemble size and a tree depth. We train models with learning rates $0.1, 0.2, 0.3, \ldots, 0.9, 1.0$, pick the best value $r$, and then again train models with learning rates $r-0.08, r-0.06, \ldots, r+0.06, r+0.08$. We use the value with the best predictive performance on the test set.
The number of trees and tree depth parameters are either varied (stress test and random verification tasks experiment), or provided in the main paper (robustness experiment and the two use cases). 

In the robustness experiment, we used a timeout of 10 seconds for both \ouralg{} and \merge{} per $\bar{b}$ calculation. The memory limit was set to 4Gb for both methods.
In the stress test experiment, we used a timeout of 120 seconds and a memory limit of 4Gb.
In the random verification task experiment, we used a timeout of 30 seconds and memory limit of 4Gb.
For the last two use cases, we used 30 and 10 seconds for the YouTube and soccer tasks respectively, with again a memory limit of 4Gb.

\subsection{Additional Results for Stress Test}

\begin{figure}
    \centering

    \textbf{Allstate}
    \vspace{0.51em}

    \small 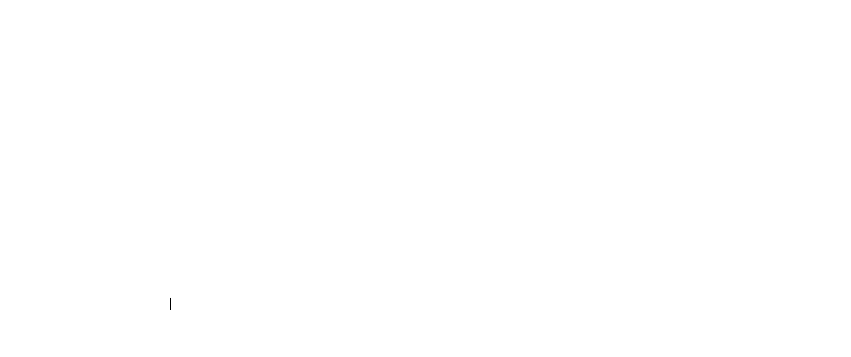

    \vspace{1em}
    \textbf{California housing}
    \vspace{0.51em}

    \small 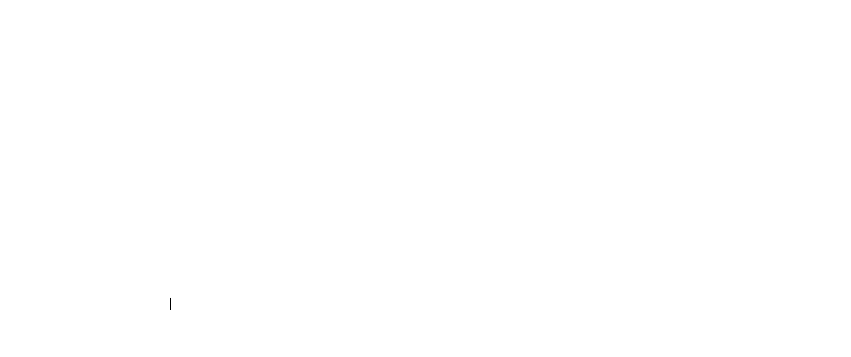

    \vspace{1em}
    \textbf{Covertype}
    \vspace{0.51em}

    \small 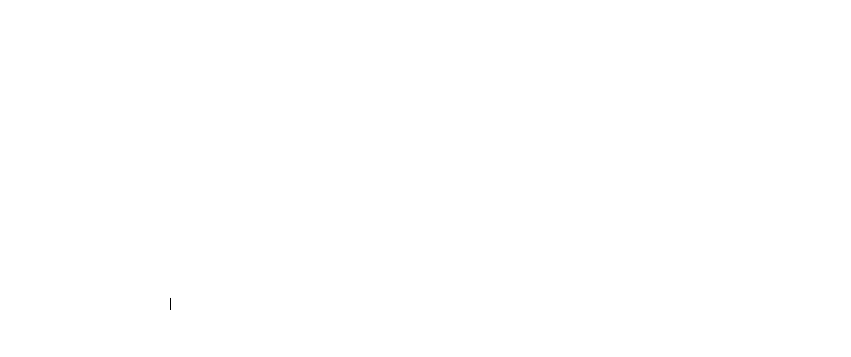

    \vspace{1em}
    \textbf{Higgs}
    \vspace{0.5em}

    \small 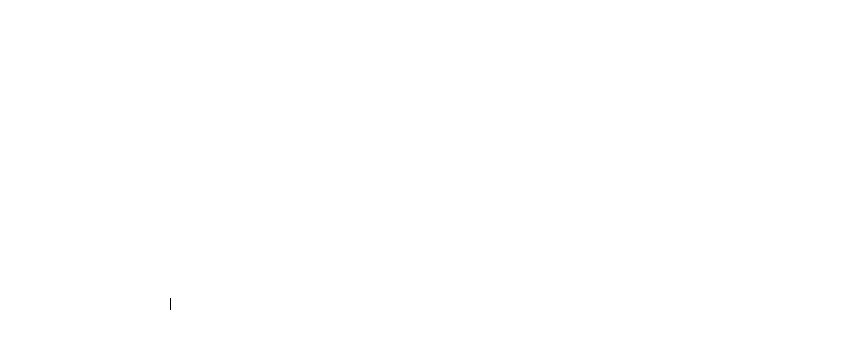

    \caption{Results for the stress test experiment for all four datasets
    The rightmost figure for Allstate shows two cases where \ouralg{} is unable to find a lower bound in time.
    This could potentially be resolved by lowering the initial $\epsilon$.}\label{fig:stresstest}
\end{figure}

Figure~\ref{fig:stresstest} shows the results for the stress test experiment for all datasets.

\subsection{Extended Results for Random Verification Task Experiment}

Tables \ref{tbl:constrainednumtrees} and \ref{tbl:constraineddepth} contain the same results as in the main paper but the results are split up per ensemble size and tree depth respectively.

Both ensemble size and tree depth positively correlate with the relative gap size. Unsurprisingly, smaller ensembles with shallower trees are easier to verify. The run time performance is fairly consistent across model sizes.

\begin{table*}
    \centering
    \footnotesize
    \newcommand{\myrndfignumtrees}[1]{\includegraphics[page=#1]{table_random_figuresnumtrees.pdf}}
\begin{tabular}{m{0.4cm}m{1.1cm}m{0.9cm}m{1.1cm}m{0.9cm}m{0.9cm}}
\toprule
   Data &  Exact \ouralg{} &  Exact \merge{} &          Gap \ouralg{} &           Gap \merge{} &                    TTB \\
\midrule
   A 25 &           85.8\% &          11.1\% &   \myrndfignumtrees{1} &   \myrndfignumtrees{2} &   \myrndfignumtrees{3} \\
   A 50 &           43.9\% &           6.1\% &   \myrndfignumtrees{4} &   \myrndfignumtrees{5} &   \myrndfignumtrees{6} \\
   A 75 &           29.4\% &           6.4\% &   \myrndfignumtrees{7} &   \myrndfignumtrees{8} &   \myrndfignumtrees{9} \\
  A 100 &           13.9\% &           4.4\% &  \myrndfignumtrees{10} &  \myrndfignumtrees{11} &  \myrndfignumtrees{12} \\
  A 150 &            9.2\% &           3.6\% &  \myrndfignumtrees{13} &  \myrndfignumtrees{14} &  \myrndfignumtrees{15} \\
  A 200 &           10.0\% &           5.3\% &  \myrndfignumtrees{16} &  \myrndfignumtrees{17} &  \myrndfignumtrees{18} \\
  CH 25 &          100.0\% &          74.2\% &  \myrndfignumtrees{19} &  \myrndfignumtrees{20} &  \myrndfignumtrees{21} \\
  CH 50 &          100.0\% &          39.2\% &  \myrndfignumtrees{22} &  \myrndfignumtrees{23} &  \myrndfignumtrees{24} \\
  CH 75 &           99.2\% &          34.7\% &  \myrndfignumtrees{25} &  \myrndfignumtrees{26} &  \myrndfignumtrees{27} \\
 CH 100 &           98.9\% &          26.7\% &  \myrndfignumtrees{28} &  \myrndfignumtrees{29} &  \myrndfignumtrees{30} \\
 CH 150 &           89.7\% &          18.1\% &  \myrndfignumtrees{31} &  \myrndfignumtrees{32} &  \myrndfignumtrees{33} \\
 CH 200 &           86.7\% &          17.5\% &  \myrndfignumtrees{34} &  \myrndfignumtrees{35} &  \myrndfignumtrees{36} \\
\bottomrule
\end{tabular}
\hspace{0.5cm}
\begin{tabular}{m{0.4cm}m{1.1cm}m{0.9cm}m{1.1cm}m{0.9cm}m{0.9cm}}
\toprule
   Data &  Exact \ouralg{} &  Exact \merge{} &          Gap \ouralg{} &           Gap \merge{} &                    TTB \\
\midrule
  CT 25 &          100.0\% &          42.5\% &  \myrndfignumtrees{37} &  \myrndfignumtrees{38} &  \myrndfignumtrees{39} \\
  CT 50 &           97.5\% &          15.3\% &  \myrndfignumtrees{40} &  \myrndfignumtrees{41} &  \myrndfignumtrees{42} \\
  CT 75 &           90.6\% &          15.6\% &  \myrndfignumtrees{43} &  \myrndfignumtrees{44} &  \myrndfignumtrees{45} \\
 CT 100 &           81.1\% &           9.7\% &  \myrndfignumtrees{46} &  \myrndfignumtrees{47} &  \myrndfignumtrees{48} \\
 CT 150 &           65.8\% &           8.3\% &  \myrndfignumtrees{49} &  \myrndfignumtrees{50} &  \myrndfignumtrees{51} \\
 CT 200 &           46.4\% &           8.1\% &  \myrndfignumtrees{52} &  \myrndfignumtrees{53} &  \myrndfignumtrees{54} \\
   H 25 &           74.7\% &          16.4\% &  \myrndfignumtrees{55} &  \myrndfignumtrees{56} &  \myrndfignumtrees{57} \\
   H 50 &           49.7\% &           8.1\% &  \myrndfignumtrees{58} &  \myrndfignumtrees{59} &  \myrndfignumtrees{60} \\
   H 75 &           35.6\% &           8.6\% &  \myrndfignumtrees{61} &  \myrndfignumtrees{62} &  \myrndfignumtrees{63} \\
  H 100 &           24.7\% &           6.1\% &  \myrndfignumtrees{64} &  \myrndfignumtrees{65} &  \myrndfignumtrees{66} \\
  H 150 &           14.4\% &           5.8\% &  \myrndfignumtrees{67} &  \myrndfignumtrees{68} &  \myrndfignumtrees{69} \\
  H 200 &           13.6\% &           6.7\% &  \myrndfignumtrees{70} &  \myrndfignumtrees{71} &  \myrndfignumtrees{72} \\
\bottomrule
\end{tabular}

    \caption{Results for randomly generated problems \textbf{per ensemble size} for Allstate (A), California housing (CH), Covertype (CT), and Higgs (H). Ensemble sizes are 25, 50, 75, 100, 150, and 200. Each row summarizes 360 random problems. The second and third column show how often a problem is solved exactly by the two methods. The \textit{gap} (($\bar{b} - \ubar{b}) / \bar{b}$) columns show the relative gap distributions. The bars show the fraction of cases where the gap is $<\!\!1\%$ (blue), up to $20\%$, $50\%$, $100\%$, and $>\!\!100\%$. A smaller gap is better.
    The \textit{TTB} histogram shows the fraction of cases where \ouralg{} is slower (first bar), or faster by a factor of up to 10, 100, 1000, and $>\!\!1000$ (next 4 green bars).
    }\label{tbl:constrainednumtrees}
\end{table*}

\begin{table*}
    \centering
    \footnotesize
    \newcommand{\myrndfigdepth}[1]{\includegraphics[page=#1]{table_random_figuresdepth.pdf}}
\begin{tabular}{m{0.4cm}m{1.1cm}m{0.9cm}m{1.1cm}m{0.9cm}m{0.9cm}}
\toprule
 Data &  Exact \ouralg{} &  Exact \merge{} &       Gap \ouralg{} &        Gap \merge{} &                 TTB \\
\midrule
  A 3 &           68.1\% &          20.8\% &   \myrndfigdepth{1} &   \myrndfigdepth{2} &   \myrndfigdepth{3} \\
  A 4 &           48.9\% &          11.4\% &   \myrndfigdepth{4} &   \myrndfigdepth{5} &   \myrndfigdepth{6} \\
  A 5 &           31.1\% &           4.2\% &   \myrndfigdepth{7} &   \myrndfigdepth{8} &   \myrndfigdepth{9} \\
  A 6 &           19.4\% &           0.6\% &  \myrndfigdepth{10} &  \myrndfigdepth{11} &  \myrndfigdepth{12} \\
  A 7 &           19.7\% &           0.0\% &  \myrndfigdepth{13} &  \myrndfigdepth{14} &  \myrndfigdepth{15} \\
  A 8 &            5.0\% &           0.0\% &  \myrndfigdepth{16} &  \myrndfigdepth{17} &  \myrndfigdepth{18} \\
 CH 3 &          100.0\% &          66.4\% &  \myrndfigdepth{19} &  \myrndfigdepth{20} &  \myrndfigdepth{21} \\
 CH 4 &           99.7\% &          45.0\% &  \myrndfigdepth{22} &  \myrndfigdepth{23} &  \myrndfigdepth{24} \\
 CH 5 &           98.9\% &          36.1\% &  \myrndfigdepth{25} &  \myrndfigdepth{26} &  \myrndfigdepth{27} \\
 CH 6 &           95.8\% &          26.4\% &  \myrndfigdepth{28} &  \myrndfigdepth{29} &  \myrndfigdepth{30} \\
 CH 7 &           89.7\% &          23.3\% &  \myrndfigdepth{31} &  \myrndfigdepth{32} &  \myrndfigdepth{33} \\
 CH 8 &           90.3\% &          13.1\% &  \myrndfigdepth{34} &  \myrndfigdepth{35} &  \myrndfigdepth{36} \\
\bottomrule
\end{tabular}
\hspace{0.5cm}
\begin{tabular}{m{0.4cm}m{1.1cm}m{0.9cm}m{1.1cm}m{0.9cm}m{0.9cm}}
\toprule
 Data &  Exact \ouralg{} &  Exact \merge{} &       Gap \ouralg{} &        Gap \merge{} &                 TTB \\
\midrule
 CT 3 &           93.3\% &          42.5\% &  \myrndfigdepth{37} &  \myrndfigdepth{38} &  \myrndfigdepth{39} \\
 CT 4 &           97.5\% &          25.8\% &  \myrndfigdepth{40} &  \myrndfigdepth{41} &  \myrndfigdepth{42} \\
 CT 5 &           92.5\% &          17.2\% &  \myrndfigdepth{43} &  \myrndfigdepth{44} &  \myrndfigdepth{45} \\
 CT 6 &           79.4\% &           7.2\% &  \myrndfigdepth{46} &  \myrndfigdepth{47} &  \myrndfigdepth{48} \\
 CT 7 &           67.8\% &           5.8\% &  \myrndfigdepth{49} &  \myrndfigdepth{50} &  \myrndfigdepth{51} \\
 CT 8 &           50.8\% &           0.8\% &  \myrndfigdepth{52} &  \myrndfigdepth{53} &  \myrndfigdepth{54} \\
  H 3 &           78.3\% &          26.9\% &  \myrndfigdepth{55} &  \myrndfigdepth{56} &  \myrndfigdepth{57} \\
  H 4 &           52.5\% &          15.0\% &  \myrndfigdepth{58} &  \myrndfigdepth{59} &  \myrndfigdepth{60} \\
  H 5 &           36.9\% &           7.8\% &  \myrndfigdepth{61} &  \myrndfigdepth{62} &  \myrndfigdepth{63} \\
  H 6 &           20.8\% &           1.4\% &  \myrndfigdepth{64} &  \myrndfigdepth{65} &  \myrndfigdepth{66} \\
  H 7 &           16.7\% &           0.6\% &  \myrndfigdepth{67} &  \myrndfigdepth{68} &  \myrndfigdepth{69} \\
  H 8 &            7.5\% &           0.0\% &  \myrndfigdepth{70} &  \myrndfigdepth{71} &  \myrndfigdepth{72} \\
\bottomrule
\end{tabular}

    \caption{Results for randomly generated problems \textbf{per tree depth} for Allstate (A), California housing (CH), Covertype (CT), and Higgs (H). Tree depths are 3, 4, 5, 6, 7, and 8. Each row summarizes 360 random problems. The second and third column show how often a problem is solved exactly by the two methods. The \textit{gap} (($\bar{b} - \ubar{b}) / \bar{b}$) columns show the relative gap distributions. The bars show the fraction of cases where the gap is $<\!\!1\%$ (blue), up to $20\%$, $50\%$, $100\%$, and $>\!\!100\%$. A smaller gap is better.
    The \textit{TTB} histogram shows the fraction of cases where \ouralg{} is slower (first bar), or faster by a factor of up to 10, 100, 1000, and $>\!\!1000$ (next 4 green bars).
    }\label{tbl:constraineddepth}
\end{table*}

\small
\bibliographystyle{aaai}
\bibliography{main}

\end{document}